\documentclass[letterpaper, 10 pt, conference]{ieeeconf}  

\IEEEoverridecommandlockouts                              

\usepackage{graphics} 
\usepackage{graphicx} 
\usepackage{mathptmx} 
\usepackage{times} 
\usepackage{amsmath} 
\usepackage{amssymb}  
\usepackage{color}
\usepackage{subfig}
\usepackage{epstopdf}
\usepackage{wrapfig}
\usepackage{adjustbox}
\usepackage{booktabs, makecell}
\usepackage{multirow}
\usepackage{times}
\usepackage{textcomp}
\usepackage[font=small]{caption}
\DeclareMathAlphabet{\mathcal}{OMS}{cmsy}{m}{n}
\usepackage[normalem]{ulem}
\usepackage{hhline}
\usepackage{diagbox}
\usepackage{balance}

\long\def\ignore#1{}

\title{\LARGE \bf
GRIP: Generative Robust Inference and Perception for Semantic\\Robot Manipulation in Adversarial Environments
}

\author{Xiaotong Chen$^{1}$, Rui Chen$^{1}$, Zhiqiang Sui$^{1}$, Zhefan Ye$^{1}$, Yanqi Liu$^{2}$,\\ R. Iris Bahar$^{2}$ and Odest Chadwicke Jenkins$^{1}$
\thanks{$^{1}$ X. Chen, R. Chen, Z. Sui, Z. Ye and O.C. Jenkins are with the Department of Electrical Engineering and Computer Science, Robotics Institute, University of Michigan, Ann Arbor, MI, USA, 48109-2121  {\tt\small [cxt|richen|zsui|zhefanye|ocj]@umich.edu}}
\thanks{$^{2}$ Y. Liu and R. Bahar are with the Department of Computer Science and the School of Engineering, Brown University, Providence, RI, USA, 02912-1910  {\tt\small [yanqi\_liu|iris\_bahar]@brown.edu }}
}

\begin{document}

\maketitle
\thispagestyle{empty}
\pagestyle{empty}
\graphicspath{{figures/}}


\begin{abstract}

Recent advancements have led to a proliferation of machine learning systems used to assist humans in a wide range of tasks.  However, we are still far from accurate, reliable, and resource-efficient operations of these systems.  For robot perception, convolutional neural networks (CNNs) for object detection and pose estimation are  recently coming into widespread use.  
However, neural networks are known to suffer from overfitting during the training process and are less robust under unforeseen conditions (which makes them  especially vulnerable to {\em adversarial scenarios}). In this work, we propose {\em Generative Robust Inference and Perception  (GRIP)} as a two-stage object detection and pose estimation system that aims to combine the relative strengths of discriminative CNNs and generative inference methods to achieve robust estimation. 
Our results show that a second stage of sample-based generative inference is able to recover from false object detections by CNNs, and produce robust estimations in adversarial conditions. We demonstrate the efficacy of {\em GRIP} robustness through comparison with state-of-the-art learning-based pose estimators and pick-and-place manipulation in dark and cluttered environments.
\end{abstract}

\section{INTRODUCTION}


Taking advantage of the renaissance in deep neural networks, machine learning has achieved great progress in object detection and segmentation and image recognition. These deep learning methods are also prevalent in robotics for problems, including manipulation in clutter~\cite{gualtieri2018pick} and learning of manipulation actions~\cite{levine2016end}. For 6D object pose estimation, learning-based Convolutional Neural Networks (CNNs) have achieved promising accuracy and real-time inference speed~\cite{xiang2017posecnn,tremblay2018deep,wang2019densefusion}. Notably, these successes rely on well-designed models and adequate training resources. The robustness and generalization capability of CNNs heavily depend on the training data, which represents a certain range of conditions that could be faced by robots. However, due to the complex and dynamic nature of the real world, robots are subject to unforeseen environmental conditions, which are not present in the training data. 
\begin{figure}[]
    \centering
    \includegraphics[width=0.8\columnwidth]{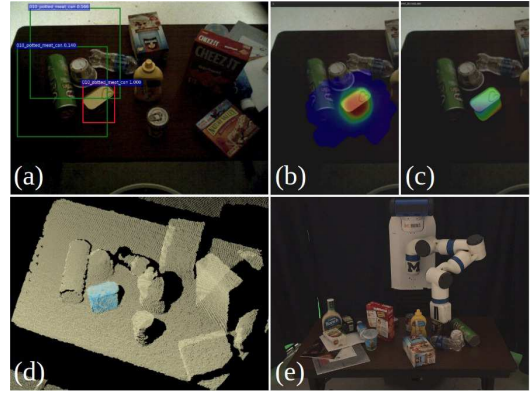}
     \caption{
     Our {\em GRIP} system perceiving and grasping an object in adversarially darkened lighting.  {\em GRIP} uses two stages of (a) PyramidCNN object detection bounding boxes with confidence score greater than 0.1 (green boxes), shown along with the ground truth (red box), and (b) sample-based generative inference. The (c) resulting estimate and (d) its localized pose (highlighted in cyan) enables (e) the Michigan Progress Fetch robot to accurately grasp the potted meat can object.
    }
    \label{fig:teaser}
\end{figure}

More specifically, CNNs recognition systems introduce vulnerability to errors (both benign and malicious) due to the effects of overfitting during the training process.  Distorted objects and/or objects captured under poor lighting conditions could be enough to defeat the recognition abilities of a CNN \cite{evtimov2017robust}. Such perception errors can lead to (potentially disastrous) outcomes for embodied systems acting in the real world.  These challenges for {\bf robust perception} become that much more challenging when an adversary can modify the environment to exploit the vulnerabilities of a CNN. For instance, in the context of object recognition for a robotic system, a possible malicious attack (through simple modifications of an environment) has the potential to drastically alter and even manipulate a robot's final behavior. Fig.~\ref{fig:teaser} shows such a robot manipulation task under dark scene.

Generative-discriminative algorithms~\cite{sui2017sum,liu2018robust} offer a promising avenue for robust perception.  Such methods combine inference by deep learning (or other discriminative techniques) with sampling and probabilistic inference models to achieve robust and adaptive perception in adversarial environments.  The value proposition for generative-discriminiative inference is to get the best out of existing approaches to computational perception and robotic manipulation while avoiding their shortcomings. We want the robustness of belief space planning~\cite{kaelbling1998planning,kaelbling2013integrated} without its computational intractability. The recall power of neural networks without excessive overfitting~\cite{levine2016end}. The efficiency of deterministic inference without its fragility to uncertainty \cite{fikes1971strips,mohan2012acquiring}. Generative-discriminative algorithms may be especially advantageous when exposed to adversarial attack, building on foundational ideas in this space~\cite{narayanan2016discriminatively,narayanan2016perch,liu2015table,joho2013nonparametric,collet2011moped}. 
Furthermore, we expect our approach will be more generally applicable to guard against broad categories of attack with a clear pathway for explanability of the resulting perceptual estimates.

In this paper, we present {\em Generative Robust Inference and Perception (GRIP)} as a two-stage method to explore generative-discriminiative inference for object recognition and pose estimation in adversarial environments. Within {\em GRIP}, we represent the first stage of inference as a CNN-based recognition distribution.  The CNN recognition distribution is used within a second stage of generative multi-hypothesis optimization.  This optimization is implemented as a particle filter with a static state process.  We show that our {\em GRIP} method produces comparable and improved performance with respect to state-of-the-art pose estimation systems (PoseCNN~\cite{xiang2017posecnn} and DOPE~\cite{tremblay2018deep}) under  \textit{adversarial scenarios} with varied lighting and cluttered occlusion. Moreover, we demonstrate the compatibility of {\em GRIP} with goal-directed sequential manipulation in object pick-and-place tasks with a Michigan Progress Fetch robot.
\section{BACKGROUND}

\begin{figure*}[!t]
    \centering
    \includegraphics[width=\textwidth]{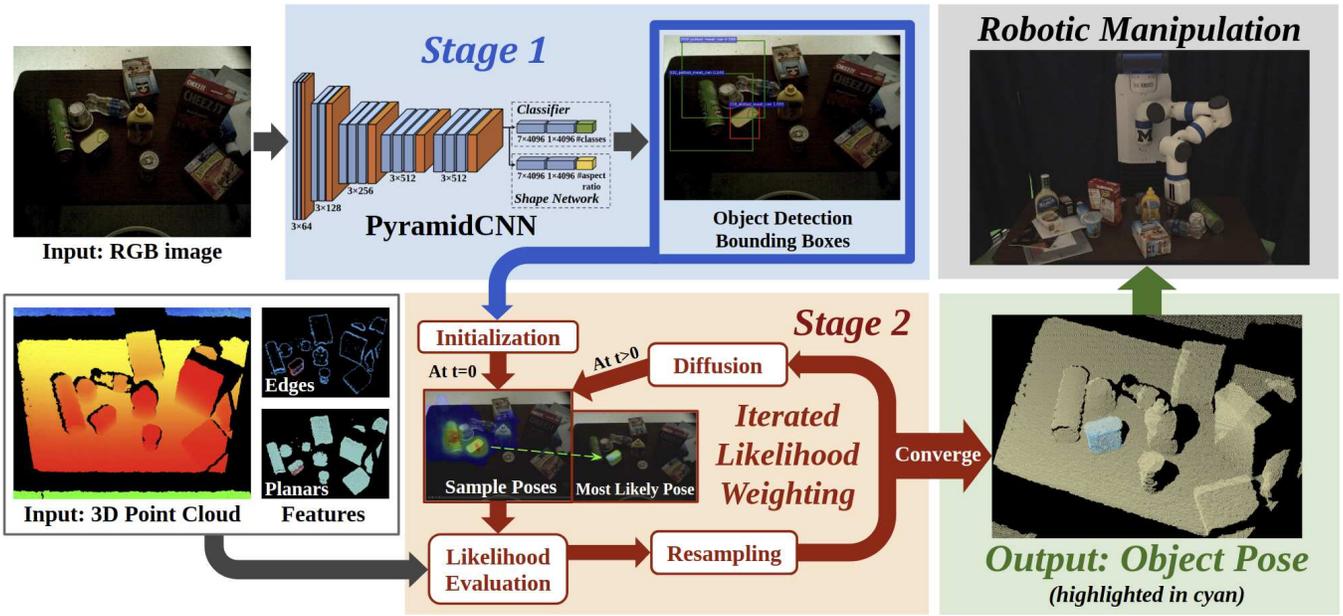}
    \caption{Overview of {\em GRIP}. The robot operating in a dark and cluttered environment is to grasp the meat can from its RGBD observation. Stage 1 takes the RGB image and generates object bounding boxes with confidence scores. Stage 2 takes the depth image and performs sample-based generative inference to estimate the pose for each object in the scene. The samples in Stage 2 are initialized according to bounding boxes from Stage 1. From this estimate, the robot performs manipulation on the meat can object.
    }
    \label{fig:diagram}
\end{figure*}

\subsection{Motivation}

To get the best of both worlds, we consider the state-of-the-art as the relative strengths and weaknesses of deep learning and generative inference for robust perception.  We are particularly interested in complementary properties of these methods for making perceptual decisions, where the weaknesses of one can be addressed by the strengths of the other.  Despite the strengths of CNNs, they have several shortcomings that leave them vulnerable to adversarial action, such as their {\em opacity} in understanding how its decisions are made, {\em fragility} for generalizing beyond overfit training examples, and {\em inflexibility} for recovering when false decisions are produced.
For these methods, Goodfellow \textit{et al.}~\cite{goodfellow2014explaining} demonstrated that adversarial examples are misclassified both in the case of different architectures or different subsets of the training data.
These weaknesses for CNNs play to the strengths of robustness for generative probabilistic inference, which are inherently: {\em explainable, general, and resilient} through the process of generating, evaluating, and maintaining a distribution of many hypotheses representing possible decisions. However, this robustness comes at the cost of computational efficiency. Probabilistic inference, in contrast to CNNs, is often computationally intractable with complexity that grows exponentially with the number of variables. 
{\em GRIP} aims to overcomes these limitations by combining the strengths of deep learning and probabilistic inference through a two-stage algorithm, illustrated in Fig.~\ref{fig:diagram} and discussed later in Section~\ref{section:methodology}.  The remainder of this background section will provide a broader overview of related existing works. 
\ignore{
This model leads us to a two-stage recognition method that considers factors from the process of detection (e.g., CNN-based convolution [17]) and process of object pose inference (e.g., Monte Carlo Localization [36]).  Estimates taken from this joint model provide a final decision about the scene to be acted upon for manipulation, while maintaining probability mass about other viable explanations of the scene. A central point of exploration for SAGE is to avoid making hard decisions and thresholds until a final detection estimate is required for recognition and the consequent robot action. For example, in the first stage, a CNN would return the evaluation of all possible bounding box regions for all object labels in a color image. In the second stage, this convolution can then be used as a factor in a generative Monte Carlo sampling process for pose estimation on depth images.  We posit that this approach will prove more robust for handling objects and scenes that have been maliciously manipulated or altered.
}

\subsection{Perception for Manipulation}

Perception is a critical step for robotic manipulation in unstructured environments. Ciocarlie \textit{et al.}~\cite{ciocarlie2014towards} proposed an architecture for reliable grasping and manipulation, where non-touching, isolated objects are estimated by clustering the surface normal of RGBD sensor data. The MOPED framework~\cite{collet2011moped} has been proposed for object detection and pose estimation using iterative clustering estimation from multi-view features. A bottom-up approach is taken in~\cite{papazov2012rigid} using RANSAC and Iterative Closest Point registration (ICP), relying solely on geometric information. Narayanan \textit{et al.}~\cite{narayanan2016discriminatively} integrated global search with discriminatively trained algorithms to balance robustness and efficiency, which works on multi-object identification, assuming known objects.

For manipulation in dense cluttered environments, ten Pas and Platt~\cite{ten2016localizing} showed success in detecting grasp affordances from 3D point clouds. In \cite{ten2017grasp}, they sample grasp pose candidates based on their geometric plausibility, from which feasible grasp poses are selected by a CNN. Regarding manipulation with known object geometry models, \cite{sui2015axiomatic, desingh2016physically, zeng2018semantic} proposed generative sampling approaches to scene estimation for object poses and physical support relations. However, these methods used object detection bounding boxes with hard thresholding as the prior for generative sampling, which might cause false negatives.

\subsection{Object Detection and Pose Estimation}

Learning-based approaches have been used as modules in object pose estimation systems, or directly built end-to-end approaches. Sui \textit{et al.}~\cite{sui2017sum} proposed a sample-based two-stage framework to sequential manipulation tasks, where object detection results are used as prior of sample initialization. Mitash \textit{et al.}~\cite{mitash2018robust} developed a two-stage approach, which ran stochastic sampling of congruent sets \cite{mellado2014super} to get object poses based on the semantic map from a segmentation network. Regarding end-to-end systems, PoseCNN \cite{xiang2017posecnn} was proposed by constructing a neural-network that learned segmentation, object 3D translation, and 3D rotation separately. This work also contributed an object dataset, called YCB-Video-Dataset, for benchmarking robotics pose estimation and manipulation approaches. DOPE \cite{tremblay2018deep} outperformed PoseCNN in estimation robustness in dark, occluded scenes by training the network on a synthetic dataset from domain-randomization and photo-realistic simulation. DenseFusion \cite{wang2019densefusion}, utilized two networks to extract RGB and depth features separately. 

In this paper, we focus on the pose estimation problem in \textit{adversarial scenarios}. Liu \textit{et al.}~\cite{liu2018robust} provided insight into handling adversarial clutter, yet provided limited evaluations of its approach or comparisons with state-of-the-art methods. We believe that the performance of CNNs relies highly on the consistency of the testing environment to the training set, and that the same is true for the two-stage methods in \cite{sui2017sum} and \cite{mitash2018robust} since they rely on high-quality CNN output from their first stages. Our main contribution is the development of a two-stage pose estimation system that is robust under adversarial scenarios and able to recover from false detections from its own first stage. 

\section{ PROBLEM FORMULATION}

Given an RGB-D observation ($Z_r$, $Z_d$) from the robot sensor and 3D geometry models of a known object set, our aim is to estimate the conditional joint distribution $P(q,b|o, Z_r ,Z_d)$ for each object class $o$, where $q$ is the six DoF object pose and $b$ is the object bounding box in the RGB image. The problem can be formulated as:
\begin{align}
\label{eq: firstexp}
& P(q,b|o, Z_r ,Z_d)\\
\label{eq: secondexp}
& =P(q|b,o,Z_r,Z_d)P(b|o, Z_r ,Z_d)\\
& =\underbrace{P(q|b,o,Z_d)}_\text{pose estimation}\underbrace{P(b|o, Z_r)}_\text{detection}
\label{eq: objdist}
\end{align} 

Equations \eqref{eq: firstexp} and \eqref{eq: secondexp} are derived using chain rule statistics and Equation \eqref{eq: objdist} represents the factoring of object detection and pose estimation. Here, we assume that pose estimation is conditionally independent of RGB observation, while object detection is conditionally independent of depth observation.

Ideally, we could use Markov Chain Monte Carlo (MCMC) \cite{hastings1970monte} to estimate the distribution of 
Equation \eqref{eq: firstexp}. However, the state space of the entire states is so large that it is intractable to directly compute. End-to-end neural network methods can also be used to calculate the distribution \cite{xiang2017posecnn,tremblay2018deep,wang2019densefusion}. 
These results place a heavy reliance on proper coverage of the input space in the training set.  This data reliance makes such methods vulnerable to unforeseen environment changes. SUM \cite{sui2017sum} implements a combination of Equation \eqref{eq: firstexp} to filter over hard detections provided by a CNN, thereby enabling it to filter out false positive CNN detections. The limitation of SUM is its inability to recover from false negatives that are eliminated from consideration in object proposal and detection stages. On the other hand, our {\em GRIP} paradigm is able to compensate for data deficiency by employing a generative sampling method in the second stage. 
\section{METHOD} \label{section:methodology}

We propose a two-stage paradigm to combine object detection and pose estimation, as shown in Fig.~\ref{fig:diagram}.
In the first stage of inference, PyramidCNN performs object detection and generates a prior distribution $P(b|o, Z_r)$ of 2D bounding boxes for each object label $o$. In the second stage, we perform generative multi-hypothesis optimization to estimate the joint distribution $P(q,b|o,Z_{(r,d)})$ for each object label $o$ using the first stage output as prior. The second stage is implemented as an iterated likelihood weighting filter \cite{Mckenna2007TrackingHM}:
\begin{align}
\underbrace{P(q_{0},b_{0}|o,Z_{(r,d)})}_\text{Sample Initialization} =&~{P(q_0|b_0)}P(b_0|o, Z_r)\label{eq:sampleInit}\\
P(q_t,b_t|o,Z_{(r,d)}) =&~\eta \underbrace{P(Z_{(r,d)}|q_t, b_t, o)}_\text{Likelihood}\underbrace{\overline{P}(q_t,b_t|o, Z_{(r,d)})}_\text{Proposal}\label{eq:pf_correction}\\
\overline{P}(q_t,b_t|o, Z_{(r,d)}) =& \int\int \underbrace{P(q_t,b_t|q_{t-1},b_{t-1})}_\text{Diffusion}\cdot\nonumber\\
~&~~~~P(q_{t-1},b_{t-1}|o,Z_{(r,d)})dq_{t-1}db_{t-1}\label{eq:pf_predict}
\end{align}
where $\eta$ is the normalizing factor. 
In 
Equation~\eqref{eq:sampleInit}, initial pose $q_0$ is generated from bounding boxes $b_0$, which are sampled from the prior distribution generated by the first stage. 
After the second stage, we get a probability distribution of pose estimation as shown in Equation~\eqref{eq: firstexp}. We consider the best estimate as the one with highest probability. Equivalently, best pose $q^{*}$ satisfies,
\begin{equation}
    q^{*},b^{*} = \underset{q,b}{\mathrm{argmax}}~P(q,b|o,Z_r,Z_d)
\end{equation}

\subsection{Object Detection}
\label{subsec:objdetect}

The goal of the first stage is to provide a probability distribution map for an object class $o$ in a given input image. To achieve this, we exploit the discriminative power of CNNs. Inspired by region proposal networks (RPN) in~\cite{ren2015faster}, our PyramidCNN serves as a proposal method for the second stage. We choose VGG-16 networks \cite{simonyan2014very} to extract features, which are directed to two fully convolutional networks (FCN)~\cite{long2015fully}: a classifier learning the object labels and a shape network learning the bounding box aspect ratios. The structure of PyramidCNN is detailed in Fig.~\ref{fig:diagram}.

The input to our networks is a pyramid of images at different scales. This enables the networks to detect objects with different sizes and appearing at various distances. Thus, the output contains a pyramid of heatmaps representing bounding boxes associated with confidence scores, positions, aspect ratios, and sizes for each object class. Different from end-to-end learning systems, we do not apply any threshold to the confidence scores in order to avoid any false negatives generated by the first stage.

\subsection{Pose Estimation}

The purpose of the second stage is to estimate the object pose by performing iterated likelihood weighting, which offers us robustness and versatility over the search space.  This is critical in our context since the manipulation task heavily depends on the accuracy of pose estimations. We expect the second stage to perform robustly even with inaccurate detection from the first stage.

\subsubsection{Initial Samples}
We use a set of weighted samples $\{q^{(i)}, w^{(i)}\}^M_{i=1}$ to represent the belief of object pose, where each 6D sample pose $q^{(i)}$ corresponds to a weight $w^{(i)}$. Given an object class $o$, its pose $q$, and the corresponding geometry model, we can render a 3D point cloud observation $\textit{\textbf{r}}$ using the z-buffer of a 3D graphics engine. Essentially, these rendered point clouds are what would be observed if the object had the hypothesized poses, which we refer to as \textit{rendered samples} hereafter. The samples are initialized according to the first stage output. As mentioned in Section \ref{subsec:objdetect}, our CNN produces a density pyramid that is essentially a list of bounding boxes with confidence scores. We perform importance sampling over the confidence scores and initialize our samples uniformly within the 3D workspaces indicated by sampled bounding boxes as shown in Equation~\eqref{eq:sampleInit}. More samples are spawned within bounding boxes with higher confidence scores.

\subsubsection{Likelihood Function}
\label{subsubsec: likelihood}
The weight of each sample is calculated by the likelihood function, which evaluates the compatibility of a sample with observations as shown in Equation~\eqref{eq:pf_correction}. The likelihood function consists of several parts, including bounding boxes weight, raw pixel-wise inlier ratio, and feature-based inlier ratio. We first define the raw pixel-wise inlier function as:
\begin{equation}
\label{eqn:inlier}
    \text{Inlier}(p, p^{'}) = \mathbf{I}\left(||p-p^{'}||_2 < \epsilon\right),
\end{equation}
\noindent where $p, p^{'}\in \mathbb{R}^3$ refer to a point in observation point cloud $\textbf{\textit{z}}$ and a point in rendered point cloud from sample pose respectively. $\mathbf{I}$ is the indicator function. A rendered point is considered as an inlier if it is within a certain sensor resolution range $\epsilon$ from an observed point. The point-wise inlier ratio of a rendered sample is then defined as:
\begin{equation}
\label{eqn:inlier_n}
    I =\frac{1}{|\textbf{\textit{r}}|}\sum_{(u,v)\in z} \text{Inlier}(\textbf{\textit{r}}_{(u, v)},\textbf{\textit{z}}_{(u, v)}),
\end{equation}
\noindent where $(u, v)$ refers to 2D image indices in the rendered sample point cloud $\textit{\textbf{r}}$ and observation point cloud $\textit{\textbf{z}}$. $|\cdot|$ refers to point cloud size.

Besides raw point-wise inliers, we extract geometry feature point clouds from both rendered samples and observation point clouds and compute feature inlier ratios. Hereby, we enhance the robustness of the likelihood function by considering contextual geometric information from 3D point clouds. This term prunes wrong poses that agree with the observation only in individual points but neglect higher-level geometric information such as depth discontinuity and sharp object surfaces. We apply feature point extraction introduced by Zhang \textit{et al.}~\cite{zhang2014loam} based on local surface smoothness:
\begin{equation}
    c_{(u, v)} = \frac{||\sum_{(u', v') \in \mathrm{N}(u, v)}\left(\textit{\textbf{p}}_{(u', v')} - \textit{\textbf{p}}_{(u, v)}\right)||}{|\mathrm{N}(u, v)|\cdot||\textit{\textbf{p}}_{(u, v)}||}
    \label{eq:c}
\end{equation}
\noindent $c_{(u, v)}$ is calculated by adding all displacement vectors from $\textit{\textbf{p}}_{(u, v)}$ to each of its neighbor points $\mathrm{N}(u, v)$. The point cloud $\textbf{\textit{p}}$ here can be either rendered sample $\textit{\textbf{r}}$ or observation $\textit{\textbf{z}}$. The value is then normalized by the size of $\mathrm{N}(u, v)$ and the length of vector $\textit{\textbf{p}}_{(u,v)}$. Intuitively, $c$ describes the depth changing rate within a certain local range, which has larger values in areas with acute depth changes and smaller values where object surfaces are consistent. We extract two features, edge points and planar points, by selecting point sets with largest and smallest $c$ values respectively. To balance feature point density in areas with different observation quality, we set a maximum number of edge points and planar points to be extracted from a certain local area. Essentially, a point at $(u, v)$ can be selected as an edge or a planar point only if its $c$ value is larger or smaller than a threshold and if the number of selected points has not exceeded the limit. We find that the algorithm is insensitive to our feature extraction parameters. Finally, we apply feature extraction on both rendered sample and scene observation point cloud to get sample features and observation features. We use the same inlier calculation in Equations \eqref{eqn:inlier} 
and \eqref{eqn:inlier_n} to calculate feature inlier ratios.

The weight $w$ of each hypothesis $q$ is defined as 
\begin{equation}
\label{eqn:weight}
    W(q) = \underbrace{\alpha_{box}w_{box} + \alpha_{b}{I_b}}_\text{network terms} + \underbrace{\alpha_{r}{I_r} + 
    \alpha_{e}I_{e} + \alpha_{p}I_{p}}_\text{geometric terms}
\end{equation}
where $w_{box}$ is the confidence score of the bounding box. $I_r$ is the ratio of pixel-wise inliers in the whole rendered sample point cloud. $I_b$ is the inlier ratio in the portion of rendered sample that is within the bounding box ($I_b$ is 0 if no rendered sample point falls into the bounding box). $I_{e}$ and $I_{p}$ are inlier ratios in sample edges and sample planars with respect to observation features.
The coefficients $\alpha_*$ represent the importance of each likelihood term and sum up to 1. The first two terms, $w_{box}$ and $I_b$, \textit{network terms}, are heavily determined by the bounding boxes and describe the consistency between pose sample and detection result. The last three terms, \textit{geometric terms}, weigh how much the current hypothesis explains itself in the scene geometry.

\subsubsection{Update Process}
To produce object pose estimations, we follow the procedure of iterated likelihood weighting by first assigning a new weight to each sample. Resampling is done with replacement according to sample weights. During the diffusion process shown in Equation~\eqref{eq:pf_predict}, each pose $q_t^{(i)}$ is diffused in the space subject to zero-mean Gaussian noises $\mathcal{N}_{T,t}(0, \sigma_{T,t}^2)$ and $\mathcal{N}_{R,t}(0, \sigma_{R,t}^2)$ with time-varying variances for translation and rotation respectively. The standard deviations $\sigma_{T,t}$ and $\sigma_{R,t}$ at iteration $t$ are decayed according to $W(q_t^{*})$, the weight of best pose estimation $q_t^{*}$ at that iteration. Bounding boxes $b_t^{(i)}$ are diffused uniformly within the image. The algorithm terminates when $W(q_t^{*})$ reaches a threshold $\bar{w}$, or the iteration limit is reached. Finally, we assume the pose weights of objects in the scene will be much higher than those for non-existing objects.

\section{EXPERIMENTS}

\subsection{Implementation}

We use PyTorch for our CNN implementation based on a VGG16 model pre-trained on ImageNet \cite{deng2009imagenet}. (more architectures are tested in \cite{liu2018robust}). The shape network branch of our CNN predicts 7 different aspect ratios. The size of a training image is 224$\times$224 and contains a single object. The aspect ratio of an object in the training image can be inferred from the width and height of the object. We apply a softmax at the end of the network to generate probability distribution of object classification and aspect ratio. We use cross entropy as the loss function in training.

\begin{figure}
    \centering
    \includegraphics[width=\columnwidth]{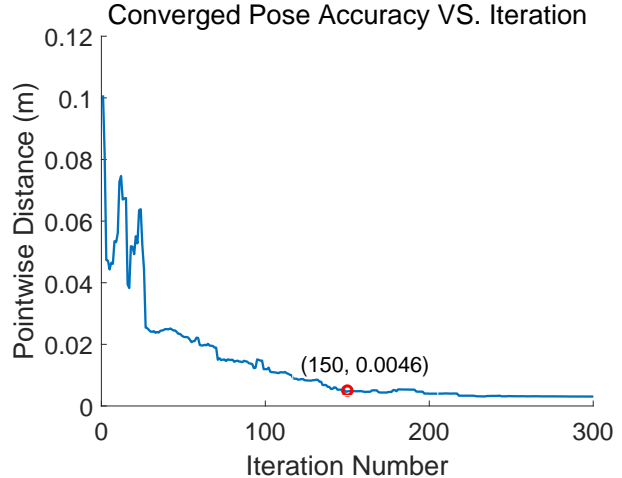}
    \caption{Plot of pose accuracy vs.\/ iteration numbers of all converged trials. The point-wise distance is calculated using ADD and ADD-S metrics \cite{xiang2017posecnn} and not used in pose estimation. After about 150 iterations, the point-wise distance reaches below 0.005m.}
    \label{fig:iteration}
\end{figure}

Our second stage pose estimation relies on the OpenGL graphics engine to render depth images with 3D geometry models and hypothesis poses on Nvidia GTX1080/RTX2070 graphics cards. During the iterated likelihood weighting process, we allocate 625 samples for each iteration and run the algorithm for 400 iterations in total, with $\epsilon$ set to 0.005m. The sample size is limited by the buffer size of our rendering engine, while the iteration limit was set since our pose estimation converges after approximately 150 iterations (see, e.g., Fig.~\ref{fig:iteration}) in less than 10s ($\sim$60ms per iteration). Point distance threshold $\epsilon$ was set to approximated distance between adjacent points in 3D point clouds.

In the feature extraction mentioned in Section~\ref{subsubsec: likelihood}, we extract up to 5 edge points and 2 planar points from each 5$\times$5 pixel non-overlap sliding window. Given a 3D point cloud $\textbf{\textit{p}}$, we consider $\textbf{\textit{p}}_{(u,v)}$ as an \textit{edge point} if $ln(c_{(u,v)})\geq-5.5$ (see Equation~\eqref{eq:c}), or a \textit{planar point} otherwise. These hyper-parameters are determined experimentally for clear indication of object boundaries as well as surfaces. The likelihood coefficients are set to $\alpha_{box}=0.1, \alpha_{b}=0.1, \alpha_{r}=0.3, \alpha_{e}=0.25, \alpha_{p}=0.25$. Through experiments, we find that the system performance is sensitive to the total category weight allocated to network terms and geometric terms, rather than the allocation within each category. If the first stage produces accurate detection evaluated by mean average precision (mAP), one can take advantage of it by allocating more weight to network terms. Otherwise, one should reduce the weight of network terms to attenuate the negative impact of underperforming first stage. Since our first stage produces low-mAP detection, we allocate only 20\% of the weight on network terms. We allocate the remaining 80\% to geometric terms since these terms are robust to adversarial scenarios and unreliable first stage detection. Further weight allocation within each category is done approximately evenly.

During diffusion, standard deviations of the Gaussian noises are decayed by a common factor $\lambda_t$, which drops exponentially from 1.0 to 0.0 as $W(q_t^{*})$ increases from 0.6 to 1.0. In other words, the standard deviations at iteration $t$ are given by $\sigma_{*,t} = \lambda_t\sigma_{*,0}$, where
\begin{equation}
\lambda_t = \mathbf{I}_{W(q_t^*)\geq0.6}\cdot\left(\frac{1-W(q_t^*)}{1-0.6}\right)^5+\mathbf{I}_{W(q_t^*)<0.6}\cdot 1
\end{equation}
Initial standard deviations are $\sigma_{T,0}=0.07m$ and $\sigma_{R,0} = 0.3rad$ for translation and rotation respectively. The threshold $\bar{w}$ for convergence is set to 0.9. 
\subsection{Dataset and Baselines}

\begin{figure}
    \centering
    \includegraphics[width=\columnwidth]{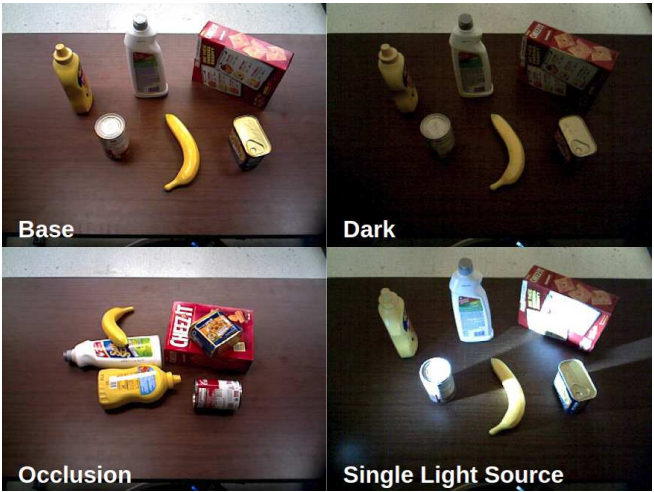}
    \caption{Testing dataset with YCB objects under adversarial settings. The base-setting data is collected with regular lighting without occlusions. The dark-setting data is collected with lights off in the room. The single lighting data is collected with a flash light. Object poses are the same in previous three settings. Data in three occlusion scenes is collected with the same objects randomly stacked.}
    \label{fig:dataset}
\end{figure}

We use the YCB video dataset \cite{xiang2017posecnn} as the training data for our first stage PyramidCNN. The YCB video dataset consists of 133,827 frames of 21 objects under normal conditions with balanced and adequate lighting but no occlusion.
To test the performance of our two-stage method with baseline methods, PoseCNN \cite{xiang2017posecnn} and DOPE \cite{tremblay2018deep}, we collect a testing dataset (i.e., adversarial YCB dataset) from 40 scenes with 15 out of 21 objects from YCB video dataset under adversarial scenarios. 
In each scene, we place 5-7 different objects on a table and collect seven frames: one in normal lighting, one in darkness, two with different single light sources, and three with different cluttered object placements (see Fig.~\ref{fig:dataset}). The dark setting and two single-lighting settings cause bias in image pixels values from the training set and thus undermine network prediction. We refer to these settings as \textit{varied lighting}
for simplicity. In addition, object clutter causes occlusions as well as natural information loss and challenges the robustness of pose estimation algorithms. All the scene images and 3D point clouds are gathered by the RGB-D sensor on our Fetch robot. Ground truth bounding boxes and 6D poses are manually labeled.

\subsection{Evaluation}
\label{sec:eval}

\begin{figure}[]
    \centering
        \includegraphics[width=\columnwidth]{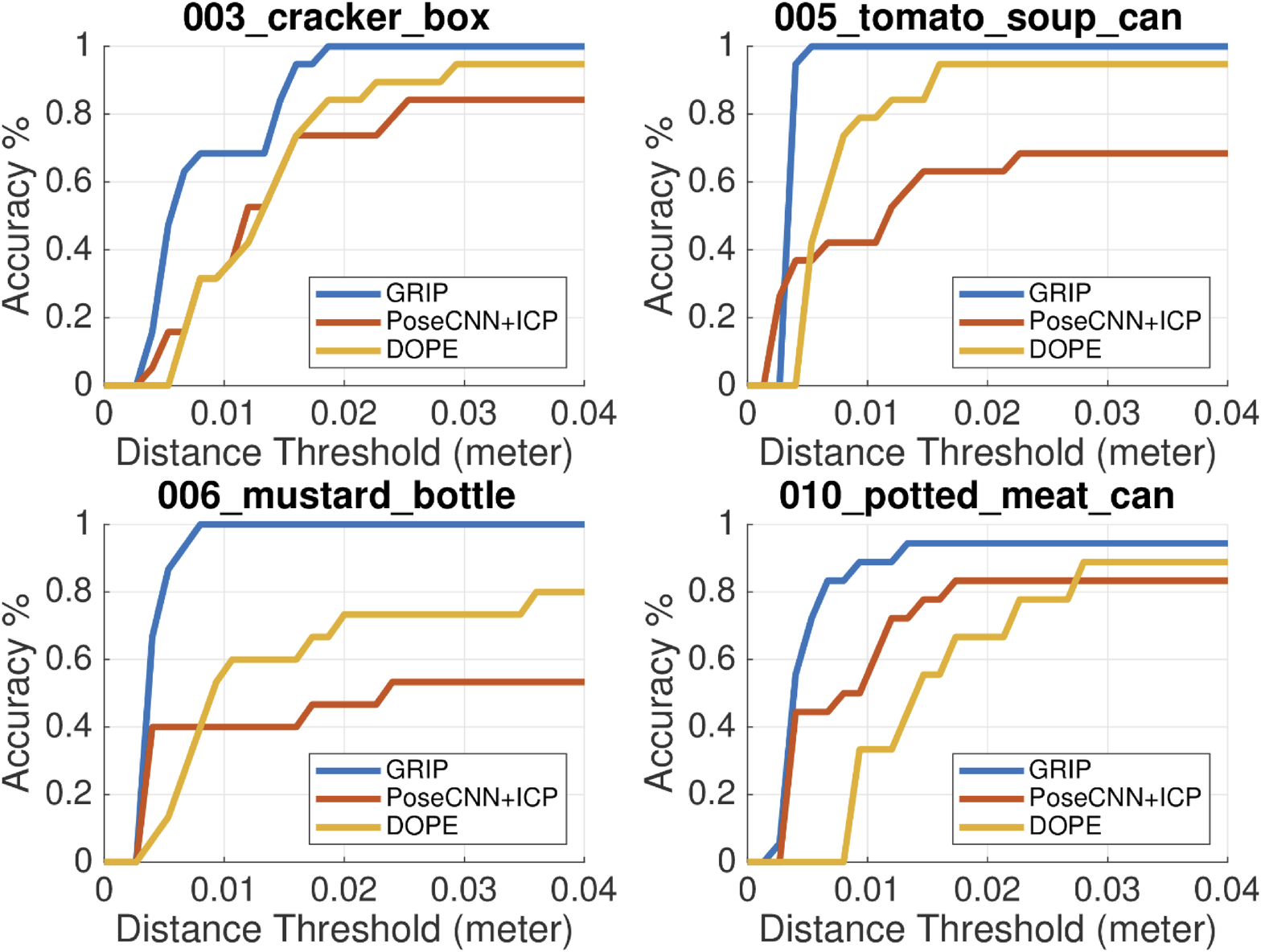}
        \vspace{0.1in}\footnotesize (a) Base settings.
        \includegraphics[width=\columnwidth]{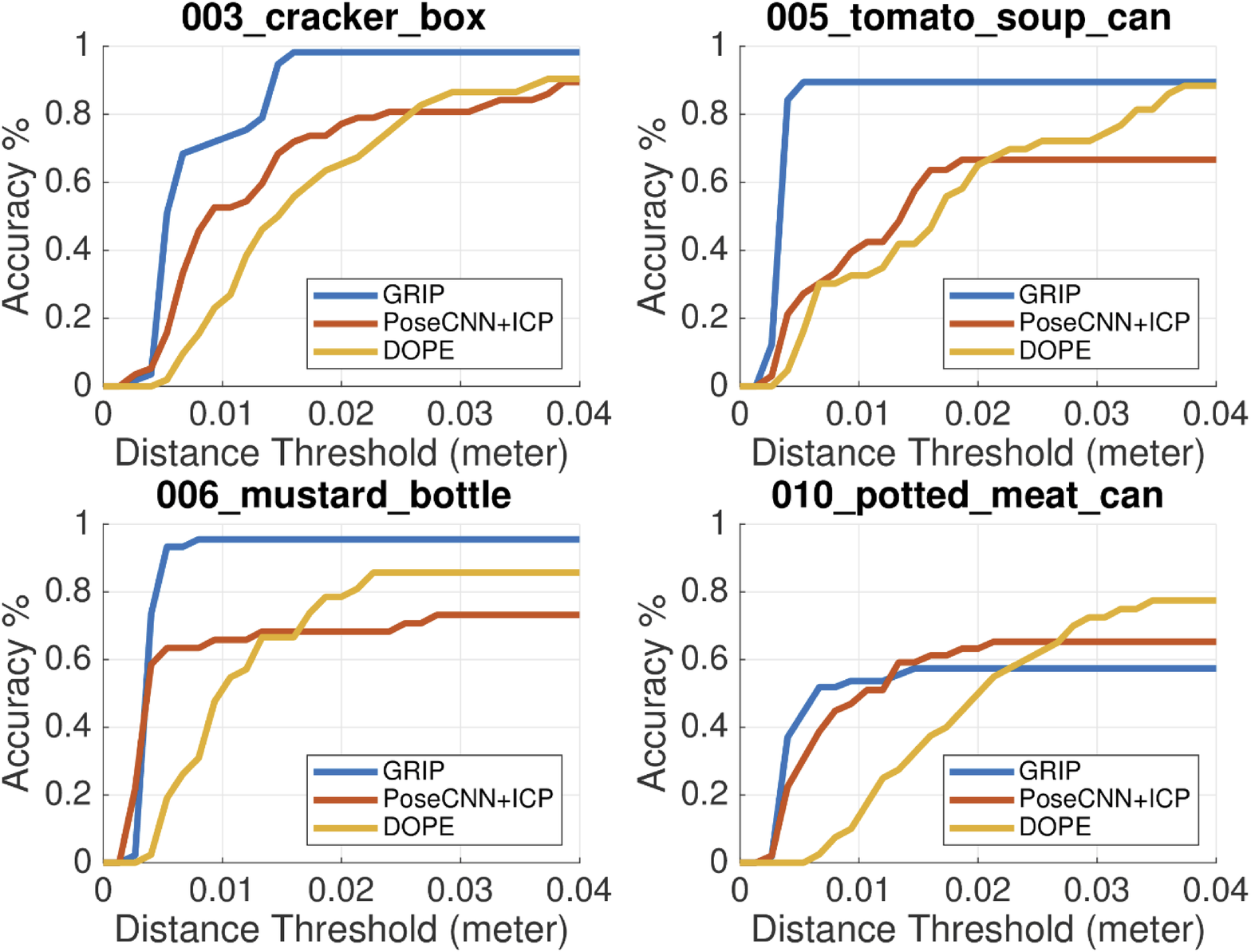}
        \vspace{0.1in}\footnotesize (b) Varied Light settings.
        \includegraphics[width=\columnwidth]{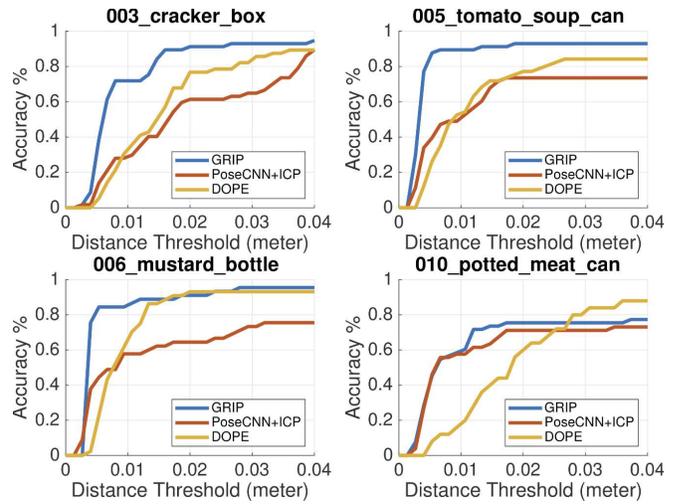}
        \vspace{0.1in}\footnotesize (c) Occlusion settings
    \caption{The comparison between DOPE, PoseCNN+ICP and our {\em GRIP}
            two-stage method on pose estimation accuracy of 4 objects mentioned in Sec. \ref{sec:eval}.}\label{fig:4objects}
\end{figure}

\begin{figure}[t]
    \centering
    	\includegraphics[width=\columnwidth]{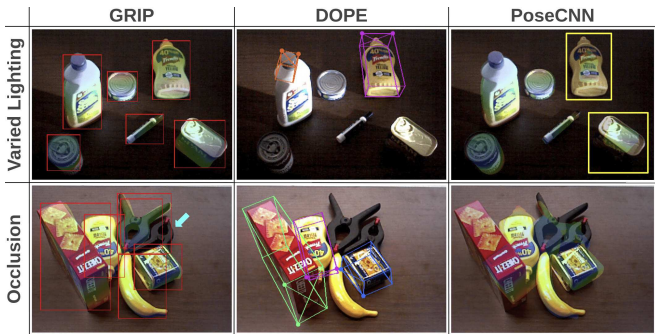}
    \caption{Comparison of {\em GRIP}, DOPE and PoseCNN under adversarial scenarios.
    In varied lighting condition, DOPE only detects 006\_mustard\_bottle correctly while PoseCNN makes inaccurate depth estimation (marked yellow). In occlusion condition, DOPE misses half of the objects while PoseCNN fails to detect object poses in clutter.
    {\em GRIP} correctly detects all objects under both scenes except 051\_large\_clamp under occlusion setting (cyan arrow) where the sampling converges to object 052\_extra\_large\_clamp because of their geometric similarity.}
    \label{fig:example_cmp}
\end{figure}

\begin{figure*}[]
    \centering
    	\includegraphics[width=\textwidth]{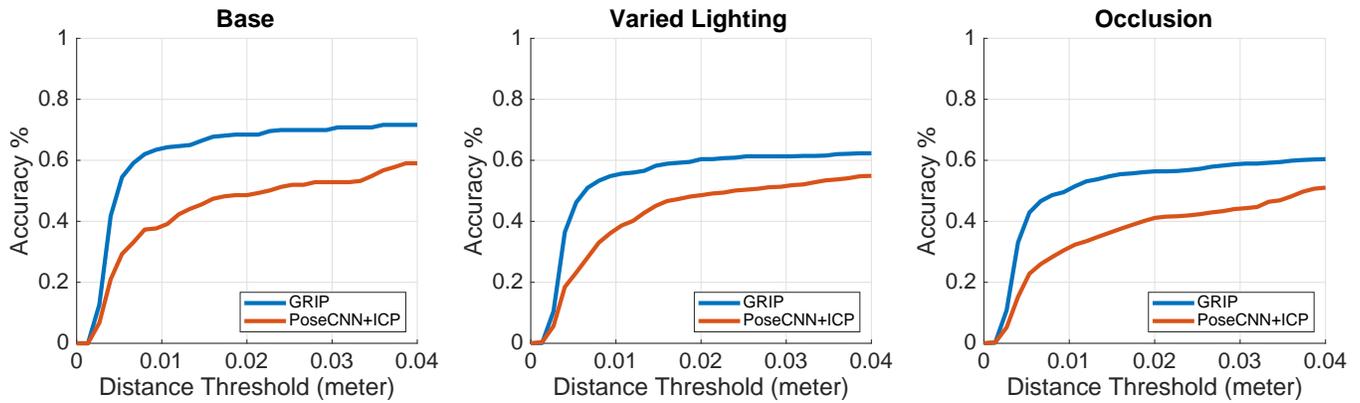}
    \caption{Overall pose estimation accuracy of 15 YCB objects using PoseCNN and our {\em GRIP}
                method.}
    \label{fig:overall}
\end{figure*}

\begin{table*}[]
    \normalsize
    \centering
    \resizebox{\textwidth}{!}{\begin{tabular}{|l|c|c|c||c|c|c||c|c|c|}
    \hline
        \multirow{2}{*}{\parbox{3.5cm}{Area Under Accuracy-Threshold Curve}} &
        \multicolumn{3}{c||}{Base} & 
        \multicolumn{3}{c||}{Varied Lighting} &
        \multicolumn{3}{c|}{Occlusions} \\ \cline{2-10}
        & DOPE & PoseCNN$^{*}$ & {\bf GRIP} & DOPE & PoseCNN$^{*}$ & {\bf GRIP} & DOPE & PoseCNN$^{*}$ & {\bf GRIP} \\
\hline 003\_cracker\_box$^{*}$ &0.6384  &0.5925  &\textbf{0.7878}  &0.5509  &0.6225  &\textbf{0.7923}  &0.5703  &0.4850  &\textbf{0.7442}  \\
\hline 005\_tomato\_soup\_can$^{*}$ &0.7691  &0.5535  &\textbf{0.9015}  &0.5326  &0.5181  &\textbf{0.8104}  &0.6372  &0.6013  &\textbf{0.8347}  \\
\hline 006\_mustard\_bottle$^{*}$ &0.5720  &0.4280  &\textbf{0.8860}  &0.6290  &0.6310  &\textbf{0.8552}  &0.7295  &0.5864  &\textbf{0.8208}  \\
\hline 007\_tuna\_fish\_can$^{*}$ &\diagbox{}{}  &0.3763  &\textbf{0.7670}  &\diagbox{}{}  &0.3616  &\textbf{0.7849}  &\diagbox{}{}  &0.3915  &\textbf{0.6220}  \\
\hline 010\_potted\_meat\_can$^{*}$ &0.5556  &0.6756  &\textbf{0.8226}  &0.4347  &\textbf{0.5273}  &0.5006  &0.5045  &0.5962  &\textbf{0.6342}  \\
\hline 011\_banana &\diagbox{}{}  &0.3922  &\textbf{0.5467}  &\diagbox{}{}  &0.3449  &\textbf{0.5591}  &\diagbox{}{}  &0.2137  &\textbf{0.2750}  \\
\hline 019\_pitcher\_base &\diagbox{}{}  &\textbf{0.2442}  &0.1774  &\diagbox{}{}  &\textbf{0.2490}  &0.1659  &\diagbox{}{}  &\textbf{0.3341}  &0.1874  \\
\hline 021\_bleach\_cleanser &\diagbox{}{}  &0.3302  &\textbf{0.5671}  &\diagbox{}{}  &0.3111  &\textbf{0.5523}  &\diagbox{}{}  &0.2204  &\textbf{0.4635}  \\
\hline 024\_bowl$^{*}$ &\diagbox{}{}  &0.3190  &\textbf{0.8674}  &\diagbox{}{}  &0.3397  &\textbf{0.7109}  &\diagbox{}{}  &0.2345  &\textbf{0.6185}  \\
\hline 025\_mug &\diagbox{}{}  &\textbf{0.3491}  &0.2201  &\diagbox{}{}  &\textbf{0.3176}  &0.2170  &\diagbox{}{}  &\textbf{0.2216}  &0.2094  \\
\hline 037\_scissors &\diagbox{}{}  &\textbf{0.5450}  &0.3192  &\diagbox{}{}  &\textbf{0.5812}  &0.1647  &\diagbox{}{}  &\textbf{0.3548}  &0.1783  \\
\hline 040\_large\_marker$^{*}$ &\diagbox{}{}  &0.2071  &\textbf{0.7537}  &\diagbox{}{}  &0.4094  &\textbf{0.6750}  &\diagbox{}{}  &0.2736  &\textbf{0.6711}  \\
\hline 051\_large\_clamp &\diagbox{}{}  &0.2405  &\textbf{0.4927}  &\diagbox{}{}  &\textbf{0.2061}  &0.1642  &\diagbox{}{}  &0.0645  &\textbf{0.2551}  \\
\hline 052\_extra\_large\_clamp &\diagbox{}{}  &\textbf{0.4000}  &0.1742  &\diagbox{}{}  &0.1460  &\textbf{0.1742}  &\diagbox{}{}  &\textbf{0.2147}  &0.1441  \\
\hline 061\_foam\_brick$^{*}$ &\diagbox{}{}  &0.8094  &\textbf{0.8333}  &\diagbox{}{}  &0.6380  &\textbf{0.8011}  &\diagbox{}{}  &0.5419  &\textbf{0.8297}  \\
\hline Overall &\diagbox{}{}  &0.4308  &\textbf{0.6078}  &\diagbox{}{}  &0.4136  &\textbf{0.5285}  &\diagbox{}{}  &0.3556  &\textbf{0.4992}  \\
    \hline
    \end{tabular}}
    \caption{Overall Performance (Area Under accuracy-threshold Curve) of 15 YCB Objects on DOPE, PoseCNN with ICP and our {\em GRIP} method. Symmetric objects are marked with stars and evaluated using ADD-S; asymmetric objects are evaluated using ADD.}
    \label{tab:overall}
\end{table*}

\subsubsection{Comparing accuracy with PoseCNN and DOPE with 4 YCB objects}
We compare our pose estimation accuracy with PoseCNN (with ICP refinement) and DOPE on the YCB dataset under different adversarial settings. We use pre-trained models from the authors' Github page for PoseCNN\footnote{https://rse-lab.cs.washington.edu/projects/posecnn/} and DOPE\footnote{https://github.com/NVlabs/Deep\_Object\_Pose} and train our first stage PyramidCNN using 2500 frames from the original YCB video dataset.
Since DOPE is trained with 5 of 21 objects from the YCB Video Dataset, we first compare all three methods on 4 of them: 003\_cracker\_box, 005\_tomato\_soup\_can, 006\_mustard\_bottle and 010\_potted\_meat\_can. The fifth object, 004\_sugar\_box, was unavailable from the market when this experiment was set up. We use ADD and ADD-S metrics \cite{xiang2017posecnn} to calculate pose error for asymmetric and symmetric objects (marked with asterisks in Table. \ref{tab:overall}) respectively. In manipulation tasks, the bearable pose estimation error is bounded by the clearance that objects have when placed in the robot end effector. Based on the sizes of Fetch robot gripper and target objects, we choose 0.04m as the maximum error tolerance. We then plot accuracy-threshold curves within a range of [0.00m, 0.04m] in Fig.~\ref{fig:4objects} and calculate AUC (Area Under accuracy-threshold Curve) as the evaluation metric. {\em GRIP} outperforms the other two methods under most error thresholds, especially lower ones, and thereby facilitates robotic manipulation tasks.

See Fig.~\ref{fig:example_cmp} for a qualitative comparison of all three methods with different adversarial settings.

\subsubsection{Comparing accuracy with PoseCNN with 15 YCB objects}
Next, we perform an extensive comparison of our method with PoseCNN (with ICP) on 15 of the 21 YCB objects.
Table~\ref{tab:overall} and Fig.~\ref{fig:overall} show our overall results and detailed accuracy evaluations for each object.

{\em GRIP} outperforms PoseCNN+ICP for most objects under all 3 settings. All methods have worse performances under varied lighting and occlusions as opposed to the basic setting. We can infer the strengths and weaknesses of each method from its performance variance among different objects. For example, PoseCNN with ICP  performs better on symmetric objects such as 003\_cracker\_box and 061\_foam\_brick as opposed to others such as 021\_bleach\_cleanser. Symmetric objects contain repetitive features which are more likely to be captured by learning-based systems. {\em GRIP} performs better on objects that are well recognizable under a depth camera. Large and compact objects such as 006\_mustard\_bottle and 024\_bowl naturally generate dense and continuous 3D point cloud observations that effectively capture their geometry. Objects with thin or articulated parts, such as 037\_scissors, 052\_extra\_large\_clamp, and 025\_mug, produce sparse point clouds around their handle-like parts that do not effectively reveal the scene geometry, especially object orientations. Hence, our {\em GRIP} algorithm best suits scenarios where rich depth sensory data are available due to detectable object dimensions and surface materials or high-definition depth sensors. Finally, distinguishing near-identical objects remains challenging. For instance, 051\_large\_clamp and 052\_extra\_large\_clamp have identical colors and shapes and differ only insignificantly in sizes. This results in poor estimation accuracy by all methods.

\subsection{Robotic Manipulation}

{\em GRIP} has been successfully used as the perception module in real robot manipulation tasks, such as a grocery packing task shown in the video\footnote{https://www.youtube.com/watch?v=W0KvzMhIJxo}.

\section{CONCLUSIONS}

We have introduced {\em GRIP} as a two-stage method for robust 6D object pose estimation suited to adversarial settings.  {\em GRIP} demonstrated similar and improved performance with respect to state-of-the-art neural network pose estimators considering the adversarial YCB dataset. The key insight of {\em GRIP} is to avoid hard thresholding, which introduces false positives and false negatives, until a final pose estimate is required. Avoiding hard thresholds increases the possibility of finding the real pose in adversarial environments.  In addition, a generative second stage inherently provides an avenue for explainable perception, without requiring deciphering network weights.  Also, this generative process readily extends to tracking over multiple instances of time through the inclusion of a proper process model. The results presented are also amenable to improvement due to the limited types of features considered.  These benefits come at the cost of assuming only one instance of each object is present in the scene.
For future work, we aim to investigate these limitations through exploring features amenable to robust inference with multiple object instances in greater clutter.

\balance





\bibliographystyle{unsrt}

\bibliography{ref}

\end{document}